\documentclass[runningheads]{llncs}
\usepackage{graphicx}

\usepackage{tikz}
\usepackage{comment}
\usepackage{amsmath,amssymb} 
\usepackage{color}
\usepackage{subcaption}
\usepackage{pifont}

\usepackage[accsupp]{axessibility}  

\usepackage[width=122mm,left=17mm,paperwidth=156mm,height=193mm,top=12mm,paperheight=217mm]{geometry}

\usepackage{booktabs}
\usepackage{xcolor,colortbl}
\usepackage{wrapfig}

\definecolor{Gray}{gray}{0.85}
\definecolor{LightCyan}{rgb}{0.88,1,1}

\newcolumntype{a}{>{\columncolor{Gray}}c}
\newcolumntype{b}{>{\columncolor{white}}c}

\newcommand{\eg}{\textit{e.g. }}
\newcommand{\ie}{\textit{i.e. }}

\renewcommand{\paragraph}[1]{{\vspace{1mm}\noindent \bf #1}.}

\usepackage[capitalize]{cleveref}
\crefname{section}{Sec.}{Secs.}
\Crefname{section}{Section}{Sections}
\Crefname{table}{Table}{Tables}
\crefname{table}{Tab.}{Tabs.}

\begin{document}

\title{SpOT: Spatiotemporal Modeling \\ for 3D Object Tracking}

\titlerunning{SpOT: Spatiotemporal Modeling for 3D Object Tracking}
\author{Colton Stearns\inst{1} \and Davis Rempe\inst{1} \and Jie Li\inst{2} \and Rares Ambrus \inst{2} \and Sergey Zakharov \inst{2} \and Vitor Guizilini \inst{2} \and Yanchao Yang \inst{1} \and Leonidas J. Guibas \inst{1}}
\authorrunning{C. Stearns et al.}
\institute{Stanford University \and Toyota Research Institute}
\maketitle

\begin{abstract}
3D multi-object tracking aims to uniquely and consistently identify all mobile entities through time. Despite the rich spatiotemporal information available in this setting, current 3D tracking methods primarily rely on abstracted information and limited history, \eg single-frame object bounding boxes. 
In this work, we develop a holistic representation of traffic scenes that leverages both spatial and temporal information of the actors in the scene. 
Specifically, we reformulate tracking as a spatiotemporal problem by representing tracked objects as sequences of time-stamped points and bounding boxes over a long temporal history. 
At each timestamp, we improve the location and motion estimates of our tracked objects through learned refinement over the full sequence of object history. 
By considering time and space jointly, our representation naturally encodes fundamental physical priors such as object permanence and consistency across time. 
Our spatiotemporal tracking framework achieves state-of-the-art performance on the Waymo and nuScenes benchmarks.

\end{abstract}

\section{Introduction}
3D multi-object tracking (MOT) is an essential task for modern robotic systems designed to operate in the real world. 
It is a core capability to ensure safe navigation of autonomous platforms in dynamic environments, connecting object detection with downstream tasks such as path-planning and trajectory forecasting.
In recent years, new large scale 3D scene understanding datasets of driving scenarios~\cite{nuscenes,waymo-open} have catalyzed research around 3D MOT~\cite{abmot3d,prob-mahabolis,gnn3dmot,kim2020eagermot}.
Nevertheless, establishing high-fidelity object tracks for this safety-critical application remains a challenge.
Notably, recent literature suggests that even small errors in 3D tracking can lead to significant failures in downstream tasks \cite{mtp,track-error-prop-2}.

A distinct challenge faced by 3D MOT is that of data association when using LIDAR data as the main source of observation, due to the sparse and irregular scanning patterns inherent in time-of-flight sensors designed for outdoor use. Established works in 2D use appearance-based association~\cite{zhang2021fairmot,bergmann2019tracking}, however, these cannot be directly adapted to 3D MOT. Sensor fusion methods combine camera and LIDAR in an effort to provide appearance-based cues in 3D association~\cite{prob-multimodal,gnn3dmot,kim2020eagermot}. However, this comes at the cost of additional hardware requirements and increased system complexity. 

Most of the recent works in 3D MOT from LIDAR data address the association problem by matching single-frame tracks to current detection results with close 3D proximity. Single-frame detection results are modeled as bounding boxes~\cite{abmot3d} or center-points~\cite{centerpoint} and compared to the same representation of the tracked objects from the last visible frame.
Although it touts simplicity, this strategy does not fully leverage the spatiotemporal nature of the 3D tracking problem: temporal context is often over-compressed into a simplified motion model such as a Kalman filter~\cite{abmot3d,prob-mahabolis} or a constant-velocity assumption~\cite{centerpoint}.
Moreover, these approaches largely ignore the low-level information from sensor data in favor of abstracted detection entities, making them vulnerable to crowded scenes and occlusions.

\begin{figure}[t]
    \centering
    \includegraphics[width=\textwidth]{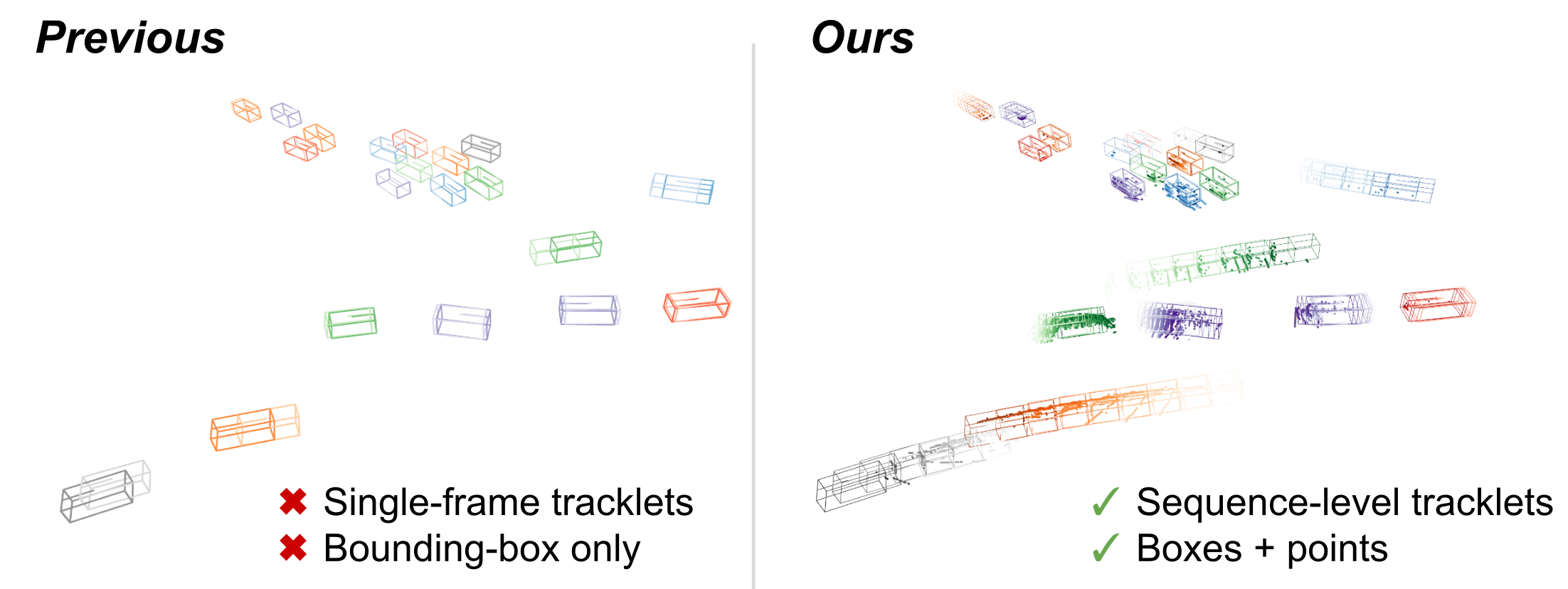}
    \caption{Previous works use a highly abstracted tracklet representation (eg. bounding boxes) and a compressed motion model (Kalman filter or constant velocity). We efficiently maintain an active history of object-level point clouds and bounding boxes for each tracklet.}
    \label{fig:teaser}
    \vspace{-3mm}
\end{figure}

However, improving spatiotemporal context by integrating scene-level LIDAR data over time is challenging due to the large quantity of sampled points along with sparse and irregular scanning patterns. Some methods aggregate LIDAR to improve 3D detection over short time horizons and in static scenes~\cite{nuscenes,hu2020you}, as well as over longer time horizons in an \textit{offline} manner~\cite{waymo-offboard}. There is also recent work for \textit{single}-object tracking that leverages low-level features in building object representations~\cite{sdf_sot,model-free}, while object-centric 4D canonical representations~\cite{caspr,oflow,dnerf} have demonstrated the power of spatiotemporal information in object reconstruction. However, these methods are restricted to object-centric datasets, require clean data (\ie low levels of noise), and run on heavy architectures that are not suitable for real time.

In this work, we propose a \textbf{spatiotemporal representation} for object tracklets (see \cref{fig:teaser}). Our method, \textbf{SpOT} (\textbf{Sp}atiotemporal \textbf{O}bject \textbf{T}racking), actively maintains the history of \textit{both} object-level point clouds and bounding boxes for each tracked object.
At each frame, new object detections are associated with these maintained past sequences, as show in \cref{fig:tracking-method}; the sequences are then updated using a novel 4D backbone to \emph{refine} the entire sequence of bounding boxes and to predict the current velocity, both of which are used to forecast the object into the next frame.
This refinement step improves the quality of bounding-box and motion estimates by ensuring spatiotemporal consistency, allowing tracklet association to benefit from low-level geometric context over a long time horizon.
We perform extensive evaluations on both nuScenes~\cite{nuscenes} and Waymo Open~\cite{waymo-open} datasets to demonstrate that maintaining and refining sequences of tracked objects has several advantages, which together enable state-of-the-art tracking performance.
Our method is particularly helpful in tracking sparse and occluded objects such as pedestrians, which can particularly benefit from temporal priors.

In summary, we contribute: (\textbf{i}) a novel tracking algorithm that leverages spatiotemporal object context by storing and updating object bounding boxes and object-level point cloud sequences,  (\textbf{ii}) a new 4D point cloud architecture for refining object tracks, and (\textbf{iii}) state-of-the-art results for 3D multi-object tracking on the standard nuScenes~\cite{nuscenes} and Waymo~\cite{waymo-open} benchmark datasets.

\section{Related Work}

\subsection{3D Object Detection on LIDAR Point Clouds}
3D detection is one of the most important modules for most 3D tracking frameworks. 
While 3D detectors from camera data have seen recent improvement~\cite{kehl2017ssd,liu2020smoke,park2021dd3d,simonelli2020demystifying}, LIDAR-based 3D detection offers much better performance, especially in driving scenes~\cite{engelcke2017vote3deep,zhou2018voxelnet,pointpillars,afdet,pv-rcnn,centerpoint}. The majority of works on LIDAR-based detection have centered around improving feature extraction from unorganized point clouds.
VoxelNet~\cite{zhou2018voxelnet} groups points by 3D voxels and extracts voxel-level features using PointNet~\cite{pointnet}. PointPillar~\cite{pointpillars} organizes point clouds in vertical columns (pillars) to achieve higher efficiency. PV-RCNN~\cite{pv-rcnn} aggregates voxel-level and point-level features to achieve better accuracy. On the other hand, CenterPoint~\cite{centerpoint} improves the 3D detector by looking at the output representation, proposing a point-based object representation at the decoding stage. While our proposed approach is not constrained to a specific input detector, we adapt CenterPoint in our experiments due to its popularity and to facilitate comparison with other state-of-the-art tracking algorithms. 

In addition to improving the 3D detector architecture, aggregating temporal information has also been shown to improve 3D detection results as it compensates for the sparsity of LIDAR sensor inputs.
nuScenes~\cite{nuscenes} devises a simple way to accumulate multiple LIDAR sweeps with motion compensation, providing a richer point cloud with an added temporal dimension. Using accumulated point clouds is shown to improve the overall performance for multiple detector architectures~\cite{pointpillars,megvii} and enables more exploration of object visibility reasoning~\cite{hu2020you}. Limited by the static scene assumption of motion compensation, temporal aggregation for the detection task is primarily done over short intervals.

Recently in offline perception, Qi et al. ~\cite{waymo-offboard} address the use of a longer time horizon as a post-processing step. After running an offline detection and tracking algorithm, Qi et al. apply a spatiotemporal sliding window refinement on each tracked object at each frame.

In our work, we explore the utilization of temporal information over a longer time horizon in the task of multi-object tracking. Distinct from Qi et al., we utilize a long-term time horizon \textit{within} our tracking algorithm, we operate in the noisier online setting, and we maintain and refine full object sequences.

\subsection{3D Multi-Object Tracking}
Thanks to the advances in 3D object detection discussed above, most state-of-the-art 3D MOT algorithms follow the \textit{tracking-by-detection} paradigm.
The performance of 3D MOT algorithms is mainly affected by three factors other than detection results: the motion prediction model, the association metric, and the life-cycle management of the tracklets.

CenterPoint~\cite{centerpoint} proposes a simple, yet effective, approach that gives reliable detection results and estimates velocities to propagate detections between sequential frames. The distance between object centers is used as the association metric. However, CenterPoint's constant velocity can be less robust to missing detections and long-term occlusions (as we demonstrate later in \cref{fig:incontiguous_seqs}).

The most popular category of 3D MOT algorithms leverages Kalman Filters to estimate
the location of tracked objects and their dynamics, providing predictions for future association. AB3DMOT~\cite{abmot3d} provides the prior baseline in this direction and leverages 3D Intersection-over-Union (IoU) as the association metric. 
Following this line, Chiu et al.~\cite{prob-mahabolis} proposes to replace 3D IoU with Mahalanobis distance to better capture the uncertainty of the tracklets.
SimpleTrack~\cite{simpletrack} conducts an analysis on different components of a tracking-by-detection pipeline and proposes corresponding enhancements to different modules. 

Other works jointly train detection and tracking in a more data-driven fashion.
FaF~\cite{luo2018fast} proposes to jointly solve detection, tracking, and prediction using a multi-frame architecture on a voxel representation. However, the architecture is hard to scale up to a long history interval.
PnPNet~\cite{pnpnet} extends the idea of FaF and proposes a more general framework with explicit tracking modeling.
Zaech et al.~\cite{zaech2022learnable} combines detection and association in a graph structure and employs neural message passing. This method provides a natural way to handle track initialization compared to heuristic methods used by previous works.

Another line of work explores feature learning for data association in sensor fusion scenarios \cite{gnn3dmot,prob-multimodal,zhang2019robust,kim2020eagermot}. In this work, we focus on the application scenario of the LIDAR sensor only.

\subsection{3D Single-Object Tracking}
Given an initial \textit{template} ground-truth bounding box of an object, the goal of single-object tracking (SOT) is to track the template object through all future frames. Unlike MOT, it is common for SOT methods to aggregate object-level information over a large temporal interval. 

Many works use a Siamese network to compare the template encoding with a surrounding region of interest. P2B \cite{p2b} uses a PointNet++ to directly propose seeds within the surrounding region and avoid an exhaustive search. BAT \cite{box-aware-feat} improves the template object representation with a box-aware coordinate space. PTTR \cite{pttr} uses cross-attention to improve feature comparison. 

Recent works propose SOT without direct supervision. Pang et. al \cite{model-free} perform template matching by optimizing hand-crafted shape and motion terms. Ye et. al \cite{sdf_sot} extend the work of Pang et. al with a deep SDF matching term.
In contrast to SOT methods, we operate in the MOT setting on sequences originally generated from an imperfect 3D detector, and we maintain object sequence histories to avoid propagating error over time.
\section{Multi-object Tracking using Sequence Refinement}
In this section, we provide an overview of our SpOT tracking pipeline and basic notation in \cref{sec:tracking-pipeline}. We then introduce our novel spatiotemporal sequence refinement module in detail in \cref{sec:method-objrep}. 

\begin{figure}[t]
    \centering
    \includegraphics[width=\textwidth]{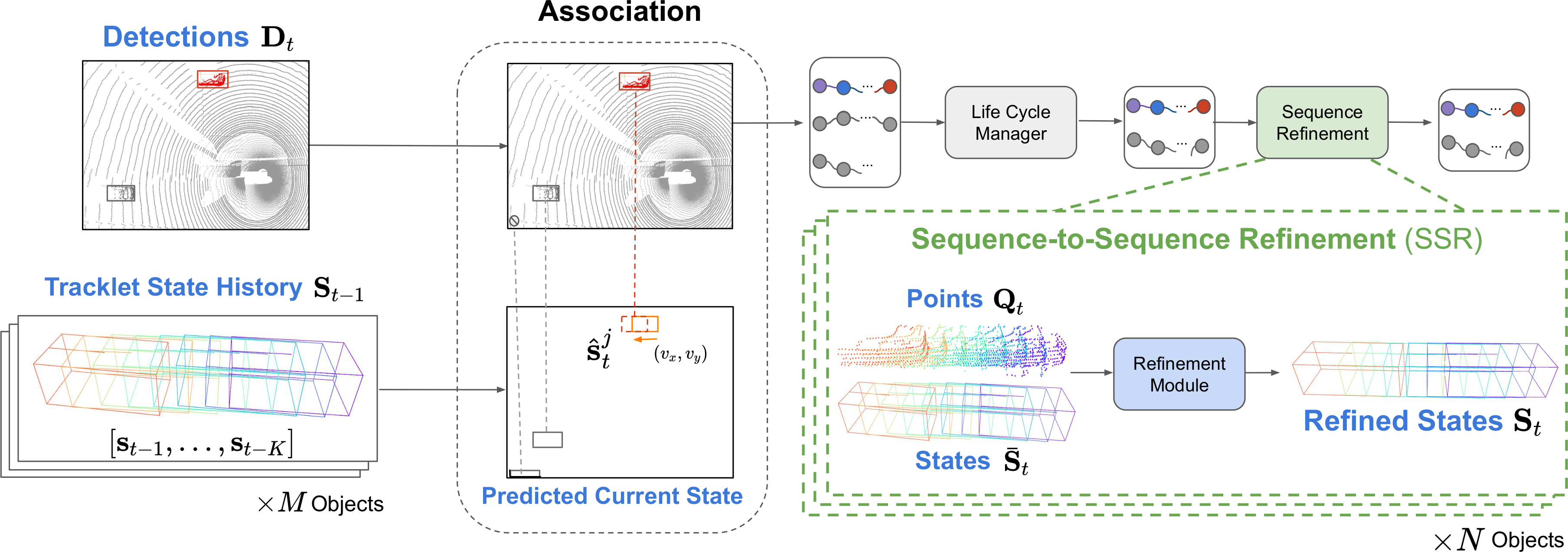}
    \caption{Tracking algorithm overview. SpOT maintains sequence-level tracklets containing both object bounding box states and object point clouds from the last $K$ timesteps. At each step, tracklets are associated to current detections using the predicted current state, and then the updated tracklets are refined by the learned SSR module to improve spatiotemporal consistency.}
    \label{fig:tracking-method}
    \vspace{-3mm}
\end{figure}

\subsection{Tracking Pipeline}\label{sec:tracking-pipeline}
%
In this work, we address the problem of 3D multi-object tracking (MOT) from LIDAR sensor input. The goal of this task is to uniquely and consistently identify every object in the scene in the form of tracklets $\mathrm{\textbf{O}}_t=\{\textbf{T}_t\}$ at each frame of input $t$. As input, the current LIDAR point cloud $\mathbf{P}_t$ is given along with detection results $\mathbf{D}_t=\{\mathbf{b}_t\}$, in the form of 7-DoF amodal bounding boxes $\textbf{b}_i=(x,y,z,l,w,h,\theta)_i$ and confidences $s_i$, from a given detector. In contrast to previous works that maintain only single-frame tracklets, we model our tracklet as $\textbf{T}_t = \{\textbf{S}_t, \textbf{Q}_t\}$ to include both a low-level history of \textit{object} points $\textbf{Q}_t$ and the corresponding sequence of detections $\mathbf{S}_t$.
%
In each tracklet, $\mathbf{S}_t$ includes the estimated state trajectory of the tracked object within a history window:

\begin{equation}
    \mathbf{S}_t = \{\textbf{s}_i=(\textbf{b},v_x,v_y,c,s)_i\},\quad t-K\leq i \leq t
\end{equation}

\noindent where $K$ is a pre-defined length of maximum history.
Tracklet state has $11$ elements and includes a 7-DoF bounding box $\textbf{b}$, a birds-eye-view velocity $(v_x, v_y)$, an object class $c$ (\eg ``car''), and a confidence score $s\in [0,1]$.
%
On the other hand, $\textbf{Q}_t$ encodes the spatiotemporal information from raw sensor observations in the form of time-stamped points:
\begin{equation}
    \textbf{Q}_t =  \{\hat{\textbf{P}}_{i}=\{(x,y,z,i)\}\}, \quad t-K\leq i \leq t
\end{equation}
where $\hat{\textbf{P}}_{i}$ is the cropped point cloud region from $\textbf{P}_i$ according to the associated detected bounding box $\textbf{b}_i$ at time $i$. We enlarge the cropping region by a factor of $1.25$ to ensure that $\hat{\textbf{P}}_{i}$ is robust through imperfect detection results.

%
As depicted in \cref{fig:tracking-method}, we propose a tracking framework that follows the tracking-by-detection paradigm while leveraging low-level sensory information. At each timestep $t$, we first predict the current tracklets $\hat{\mathbf{T}}_t$ based on the stored previous ones $\mathbf{T}_{t-1}$:
\begin{equation}
    \hat{\mathbf{T}}_t =\mathbf{Predict}(\mathbf{T}_{t-1})= \{\hat{\textbf{S}}_t,\textbf{Q}_{t-1}\}
\end{equation}
\begin{equation}
    \hat{\textbf{S}}_t = \{ \hat {\textbf{s}}_i=(x+v_x,y+v_y,z,l,w,h,\theta,v_x,v_y,c,s)_{i-1} \},\quad t-K\leq i \leq t.
\end{equation}
%
We compare the \textit{last} state of the predicted tracklets $\hat{\mathbf{s}}_t$ to off-the-shelf detection results $\mathbf{D}_t$ to arrive at associated tracklets: 

\begin{equation}
    \bar{\mathbf{T}}_t = \mathbf{Association}(\hat {\textbf{S}}_t, \textbf{D}_t, \textbf{P}_t) = \{\bar{\mathbf{S}}_t, \textbf{Q}_t\}
\end{equation}

where $\bar{\mathbf{S}}_t$ is the previous state history $\mathbf{S}_{t-1}$ concatenated with its associated detection and $\textbf{Q}_t$ is the updated spatiotemporal history.
Without loss of generality, we follow CenterPoint's~\cite{centerpoint} association strategy in our experiments. 
%
Finally, we conduct the posterior tracklet update using a novel sequence-to-sequence refinement (SSR) module:
\begin{equation}
    \textbf{S}_t = \mathbf{SSR}(\bar{\textbf{S}}_{t},\textbf{Q}_t),
\end{equation}
which provides the final updated tracklet estimation $\textbf{T}_t=\{\textbf{S}_t, \textbf{Q}_t\}$. 
In the following section, we will provide technical details of the SSR module. 
\subsection{Sequence-to-Sequence Refinement (SSR) Module}\label{sec:method-objrep}
\begin{figure}[t]
    \centering
    \includegraphics[width=1.0\textwidth]{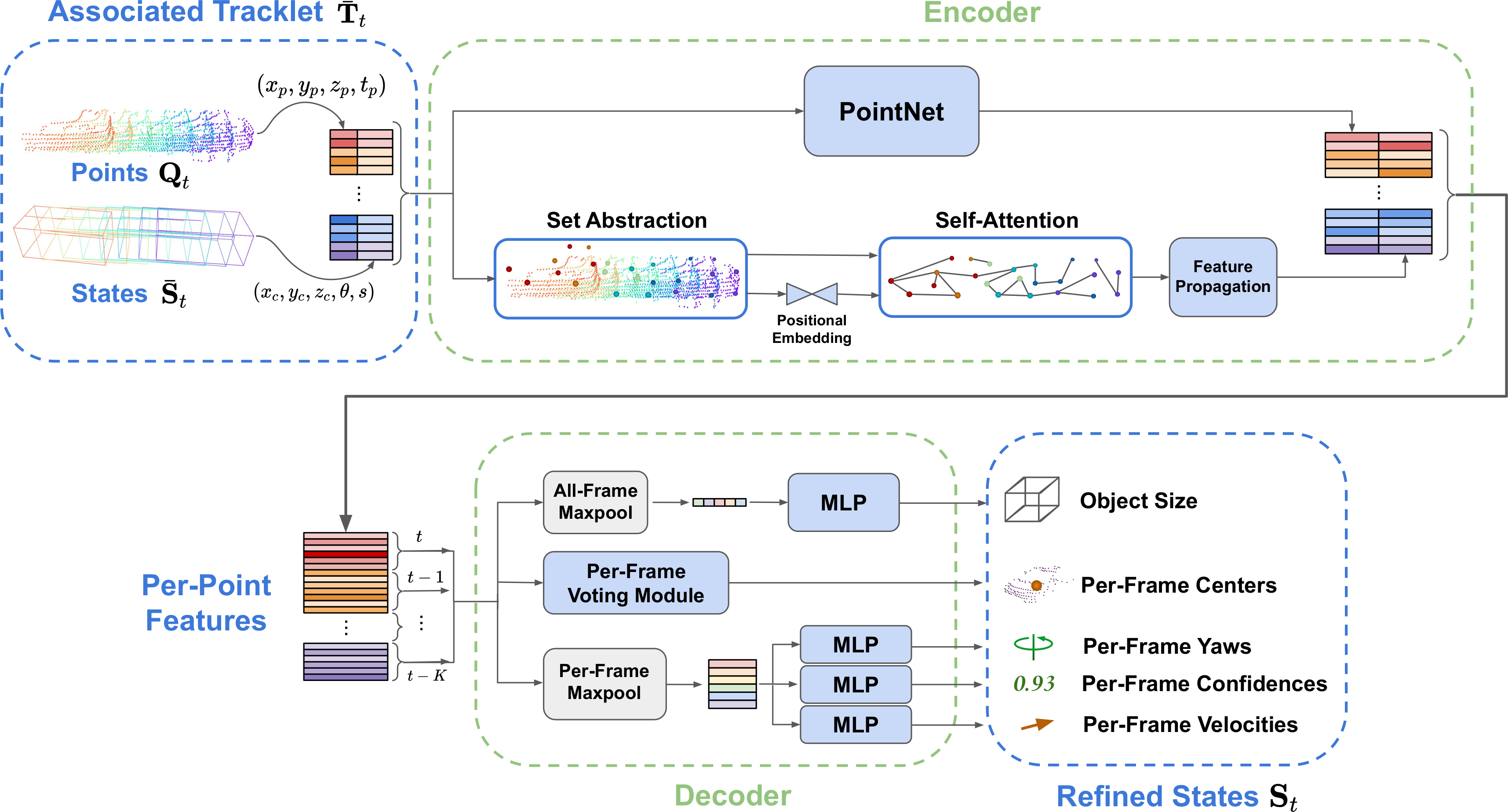}
    \caption{Architecture of the sequence-to-sequence refinement (SSR) network. Given a tracklet containing object points and bounding boxes after association to a detection, the encoder first extracts spatiotemporal features corresponding to each input point. The features are given to the decoder which predicts a refined state trajectory and velocities to be used for subsequent association.}
    \vspace{-4mm}
    \label{fig:ssr}
\end{figure}

We propose a novel algorithm to update a full tracklet history of estimated states by accounting for its spatiotemporal context, including raw sensor observations.
\cref{fig:ssr} displays our spatiotemporal sequence-to-sequence refinement (SSR) module, which takes the \textit{associated} tracklet states $\bar{\textbf{S}}_t$ and the time-stamped object point cloud segments $\textbf{Q}_t$ as input and outputs refined final tracklet states $\textbf{S}_t$. SSR first processes the sequential information with a 4D backbone to extract per-point context features. In the decoding stage, it predicts a global object size across all frames, as well as per-frame time-relevant object attributes including center, pose, velocity, and confidence. 

\paragraph{Split Self-Attention Encoder}\label{sec:split-encoder}
The top part of \cref{fig:ssr} illustrates the encoding backbone, which processes each associated tracklet independently.
Since the inputs contain two streams of information $\bar{\textbf{S}}_t$, $\textbf{Q}_t$, which are at different levels of abstraction (object vs. point), we first append the bounding-box-level information as an additional dimension to each point in $\textbf{Q}_t$. This yields a set of object-aware features:
\begin{equation}
    \mathbf{f}_p = [x_p,y_p,z_p, t_p, x_c, y_c, z_c, \sin(\theta), \cos(\theta), s],
\end{equation}
where $(x_p,y_p,z_p, t_p)$ denotes the 4D geometric point and $(x_c, y_c, z_c, \theta, s)$ is the center location, yaw, and confidence score of the corresponding bounding box at frame $t_p$.

Similar to previous works on spatiotemporal representation learning~\cite{caspr}, the encoder is a two-branch point cloud backbone as depicted in \cref{fig:ssr}. In the first branch, we apply a PointNet~\cite{pointnet} to directly encode the high-dimensional inputs into per-point features. For the second branch, we apply a novel self-attention architecture inspired by the encoder of 3Detr~\cite{3detr}. First, we apply a per-frame PointNet++~\cite{pointnet2} set abstraction layer, so that at each frame $i$ we have a sub-sampled set of anchor-point features $\{ \mathbf{a}_i^k\}_{k=1}^{A}$ where $A$ is a hyperparameter for the number of anchor points. 
For each anchor point, a 4D positional embedding is generated using a 3-layer multi-layer perceptron (MLP):
\begin{equation}
\mathbf{pos}_{\mathbf{a}^k_i} = \mathbf{MLP}(\mathbf{a}^k_i).
\end{equation}
The anchor features and positional embedding are concatenated as $[\mathbf{a}^k_i, \mathbf{pos}_{\mathbf{a}^k_i}]$ before applying four layers of self-attention across \textit{all} anchor features.
 Notably, this self-attention allows information flow across both space and time. 
 Finally, updated anchor features are propagated back to the full resolution point cloud via a feature propagation layer~\cite{pointnet2}.
 Layer normalization is applied to the features from each branch before concatenating to get the final per-point features.
 
 Each branch of the encoder uses a $256$-dim feature, yielding a concatenated $512$-dim feature at the output. Set abstraction uses $A=10$ anchor points per frame and a feature radius of $1.5 m$ for cars/vehicles and $0.6 m$ for pedestrians. Additional architectural details are provided in the supplemental material. 

\paragraph{Sequence Decoder} 
The SSR decoder outputs a refined, ordered sequence of object states $\mathbf{S}_t$ that is amenable to association in subsequent frames.
To output object state trajectories, some recent works use explicit priors on temporal continuity, such as anchor-based trajectories~\cite{multipath} or an autoregressive motion rollout \cite{adversarial-motions}.
In contrast, we choose a decoder without an explicit prior: the decoder directly predicts the ordered sequence of bounding boxes in one forward pass. 
This choice allows the model to learn temporal priors where needed through training.
Our design is motivated by the discontinuous nature of many sequences that SSR operates on, which contain identity switches, false-positives, and occlusions.

As depicted in the bottom portion of \cref{fig:ssr}, given an encoded set of per-point spatiotemporal features, we group features by their time of acquisition (\ie by frame). We pass our time-grouped point features into 5 decoding heads. The first decoding head performs a max-pool on the entire feature set to regress a single object size ($l,w,h$), which is used for every output frame. The second head applies a voting module~\cite{votenet} to each set of time-grouped features; this outputs per-timestep object center predictions ($x_c,y_c,z_c$). The remaining heads perform a max-pool on each set of time-grouped features to obtain a single feature per timestep. This feature is passed through 2-layer MLPs to regress a yaw, confidence, and velocity ($\theta,s,v_x,v_y$) for each frame. 

\paragraph{Training Losses}\label{sec:training-losses}
Our sequence refinement module balances two loss terms: a bounding box loss and a confidence-score loss. Our total loss is as follows:
\begin{equation}
    L = w_{\text{conf}}{L_{\text{conf}}} + L_{\text{box}},
\end{equation}
where $w_{\text{conf}}$ is a hyperparameter that balances the two losses.

We formulate our bounding box loss similar to standard 3D detection works \cite{centerpoint,point-lstm,pointpillars,waymo-offboard,second}. We apply an L1 loss on 3D box center $[x,y,z]$. We apply a cross-entropy loss on the predicted size bin and an L1 loss on the predicted size residual. We apply an L1 loss on the polar angle representation $\sin(\theta), \cos(\theta)$. This yields a bounding box loss of:
\begin{equation}
    L_{\text{box}} = w_{\text{c}} L_{\text{c}} + w_{\theta} L_{\theta} + w_{\text{vel}} L_{\text{vel}} + w_{\text{wlh-cls}} L_{\text{wlh-cls}} + w_{\text{wlh-res}} L_{\text{wlh-res}},
\end{equation}
where $w_{\text{c}}$, $w_{\theta}$, $w_{\text{vel}}$, $w_{\text{wlh-cls}}$, and $w_{\text{wlh-res}}$ balance losses for the bounding box center, yaw, velocity, size-bin, and size-residual, respectively.

We desire our prediction's confidence to match its quality. To achieve this, we set a target confidence score $\Bar{s}$ in a manner proportional to the accuracy of the bounding-box estimate. Concretely, if a bounding box is not close to a ground-truth object, we assign the target confidence to 0. Otherwise, we follow \cite{temperature-confidence} and assign the target confidence to be proportional to the L2 distance from the closest ground-truth object as $\Bar{s} = e^{-\alpha \mathbf{b_{err}}}$, where $\alpha$ is a temperature hyperparameter and $\mathbf{b_{err}}$ is the L2 box center error. Our confidence loss is then a binary cross-entropy loss, $L_{\text{conf}} = \text{BCE}(s, \bar{s})$.

\paragraph{Training Data}
During online tracking, the refinement module must robustly handle noisy inputs from the 3D detector, which may contain false-positives, identity switches, occlusions, and more.
Therefore, we must use a set of suitable training sequences that faithfully capture these challenging test-time phenomena.
To achieve this, we use the outputs of previous tracking methods to generate object tracks that are used as training sequences. 
For all experiments, we generate data using the CenterPoint~\cite{centerpoint} tracker with varied track-birth confidence threshold, $c_{thresh} \in \{0.0, 0.3, 0.45, 0.6\}$, and varied track-kill age, $t_{kill} \in \{1, 2, 3\}$.
We additionally augment these tracks with transformations, noise, and random frame dropping. Our final training set averages 750k sequences per object class. These augmentation methods are detailed in the supplementary material.

\section{Experimental Evaluation}
We evaluate our sequence-based tracking on the nuScenes~\cite{nuscenes} and Waymo Open~\cite{waymo-open} benchmarks. In this section, we start with an overview of the datasets and metrics in \cref{sec:dataset} and a discussion of implementation details in \cref{sec:implementation-details}. In \cref{sec:sota-tracking}, we evaluate our model performance on multi-object tracking, in \cref{sec:performance-analysis} provide ablation analyses on key design choices, and \cref{sec:runtime-analysis} evaluates runtime efficiency.

\subsection{Datasets and Evaluation Metrics}\label{sec:dataset}

\paragraph{nuScenes Dataset}
The nuScenes dataset contains 1000 sequences of driving data, each 20 seconds in length. 
32-beam LIDAR data is provided at 20Hz, but 3D labels are only given at 2Hz.
The relatively sparse LIDAR data and low temporal sampling rate make our proposed method particularly suitable for data like that in nuScenes, where leveraging spatiotemporal history provides much-needed additional context.
We follow the official nuScenes benchmark protocol for tracking, which uses the AMOTA and AMOTP metrics \cite{abmot3d}. For a thorough definition of AMOTA and AMOTP, we refer the reader to the supplementary material.
We evaluate on the two most observed classes: car and pedestrian.

\paragraph{Waymo Open Dataset}
The Waymo Open Dataset~\cite{waymo-open} contains 1150 sequences, each with 20 seconds of contiguous driving data. 
Different from nuScenes, the Waymo Open Dataset provides sensor data for four short-range LIDARs and one long-range LIDAR; each LIDAR is sampled at 10Hz, and the long-range LIDAR is significantly denser than the nuScenes 32-beam device.
Furthermore, 3D labels are provided for every frame at 10Hz.
We follow official Waymo Open Dataset benchmark protocol, which uses MOTA and MOTP~\cite{clear-mot} to evaluate tracking. For a thorough definition of MOTA and MOTP, we refer the reader to the supplementary material.
Again, we evaluate on the two most observed classes: vehicle and pedestrian.

Note that AMOTA averages MOTA at different recall thresholds. In our experiments, different sets of parameters were used between nuScenes and Waymo. For nuScenes, lower threshold were used to balance the recall.
\subsection{Implementation Details}\label{sec:implementation-details}
In this section, we highlight the most important implementation details, and refer the reader to the supplementary material for additional information.

\paragraph{SSR Training details}
We train a different network for each object class using sequences of length $K=40$ for nuScenes and $K=15$ for Waymo (we investigate the effect sequence length has on tracking performance in \cref{tab:sequence-length}).
To improve robustness and mimic test time when stored tracklets are refined iteratively each time a new frame is observed, during training we refine a sequence by a random number of times before backpropagation, \ie the network sees its own output as input.
In practice, we find that \textit{not} refining bounding-box size is beneficial for Waymo vehicles, due to the large variance in sizes. 

\begin{table}[b]
\centering
\begin{tabular}{l @{\hskip 6pt} c @{\hskip 6pt} c @{\hskip 12pt} c @{\hskip 6pt} c}
\toprule
 & \multicolumn{2}{c}{\textit{Car}} & \multicolumn{2}{c}{\textit{Pedestrian}} \\
 Method  & AMOTA$\uparrow$ & AMOTP$\downarrow$ & AMOTA$\uparrow$ & AMOTP$\downarrow$ \\
\midrule
  AB3DMOT \cite{abmot3d} & 72.5 & 0.638 & 58.1 & 0.769 \\ 
  Centerpoint \cite{centerpoint} & 84.2 & \textbf{0.380} & 77.3 & 0.392 \\
  ProbabalisticTracking \cite{prob-mahabolis} & 84.2 & -- & 75.2 & -- \\
  MultimodalTracking \cite{prob-multimodal} & 84.3 & -- & 76.6 & -- \\
  SimpleTrack-2Hz* \cite{simpletrack} & 83.8 & 0.396 & 79.4 & 0.418 \\
  \midrule
  SpOT-No-SSR (Ours) & 84.5 & \textbf{0.380} & 81.1 & 0.391\\
  SpOT (Ours) & \textbf{85.1} & 0.390 & \textbf{82.5} & \textbf{0.386} \\
\bottomrule
\end{tabular}
\vspace{3mm}
\caption{Tracking performance on the nuScenes dataset validation split. An asterisk* denotes a preprint.}
\label{tab:nuscenes-tracking-results}
\end{table}

\paragraph{Test-Time Tracking Details}
Similar to prior works~\cite{prob-multimodal,current-gnn,simpletrack}, we use off-the-shelf detections from CenterPoint~\cite{centerpoint} as input at each frame of tracking. 
For nuScenes, CenterPoint provides detections at 2Hz so we upsample to 20Hz by backtracking the estimated velocities to match the LIDAR sampling rate.
CenterPoint detections are pre-processed with a birds-eye-view non-maximal-suppression using thresholds of $0.3$ IoU for nuScenes and $0.5$ IoU for Waymo.

Detections are associated with the last frame of object tracklets using a greedy bipartite matching algorithm over L2 center-distances that uses detection confidence~\cite{centerpoint}.
We set a maximum matching distance of $1m$ and $4m$ for nuScenes pedestrians and cars, respectively, and $0.4m$ and $0.8m$ Waymo Open pedestrians and vehicles. 
We use a track-birth confidence threshold of 0.0 for nuScenes. For Waymo Open, this is 0.6 for pedestrians and 0.7 for vehicles. 
The track-kill age is 3 for both datasets. 
For nuScenes, we start refining tracklets at a minimum age of 30 frames with a maximum temporal context of 40 frames.
For Waymo, the minimum refinement age is 5 for vehicles and 2 for pedestrians and the maximum temporal context is 10 frames.

As discussed in \cref{sec:tracking-pipeline} the tracklet state trajectory $\mathbf{S}_t$ stores the history of bounding boxes for each object.
On nuScenes, these boxes are the output of our refinement network such that all boxes continue to be refined at each new frame.
On Waymo, we instead directly store the given CenterPoint detections.

\begin{table}[t]
    \centering
    \begin{tabular}{l @{\hskip 12pt} c @{\hskip 9pt} c @{\hskip 9pt} c @{\hskip 9pt} c}
        \toprule
         Method & MOTA$\uparrow$ & FP\%$\downarrow$ & Miss\%$\downarrow$ & Mismatch\%$\downarrow$ \\
         \midrule
         & \multicolumn{4}{c}{\textit{Vehicle}} \\
         AB3DMOT \cite{abmot3d} & 55.7 & -- & -- & 0.40 \\
         CenterPoint \cite{centerpoint} & 55.1 & 10.8 & 33.9 & 0.26 \\
         SimpleTrack* \cite{simpletrack} & \textbf{56.1} & \textbf{10.4} & 33.4 & \textbf{0.08} \\
         SpOT-No-SSR (Ours) & 55.1 & 10.8 & 33.9 & 0.21 \\
         SpOT (Ours) & 55.7 & 11.0 & \textbf{33.2} & 0.18 \\
         \midrule
         &  \multicolumn{4}{c}{\textit{Pedestrian}}\\
         AB3DMOT \cite{abmot3d} & 52.2 & -- & -- & 2.74 \\
         CenterPoint \cite{centerpoint} & 54.9 & \textbf{10.0} & 34.0 & 1.13 \\
         SimpleTrack* \cite{simpletrack} & 57.8 & 10.9 & 30.9 & \textbf{0.42} \\
         SpOT-No-SSR (Ours) & 56.5 & 11.4 & 31.5 & 0.61 \\
         SpOT (Ours) & \textbf{60.5} & 11.3 & \textbf{27.6} & 0.56 \\
         \bottomrule
    \end{tabular}
    \vspace{3mm}
    \caption{Tracking performance on the Waymo Open dataset validation split. An asterisk* denotes a preprint.}
    \vspace{-5mm}
    \label{tab:waymo-tracking-results}
\end{table}

\subsection{Comparison with State-of-the-Art Tracking}\label{sec:sota-tracking}
In this section, all reported tracking results are obtained with CenterPoint detections \cite{centerpoint}.
We build our tracking pipeline based on CenterPoint and adopt NMS pre-processing to detection results as suggested in SimpleTrack~\cite{simpletrack}. We denote this version of our method as \textit{SpOT-No-SSR} and report its performance for a fair comparison.

In \cref{tab:nuscenes-tracking-results}, we compare SpOT to various tracking methods on the nuScenes dataset.
SpOT significantly outperforms all previous methods in correctly tracking objects (AMOTA) and is on-par with previous methods in estimating high-quality object tracklets (AMOTP).

\begin{figure}[t]
    \centering
    \includegraphics[width=\textwidth]{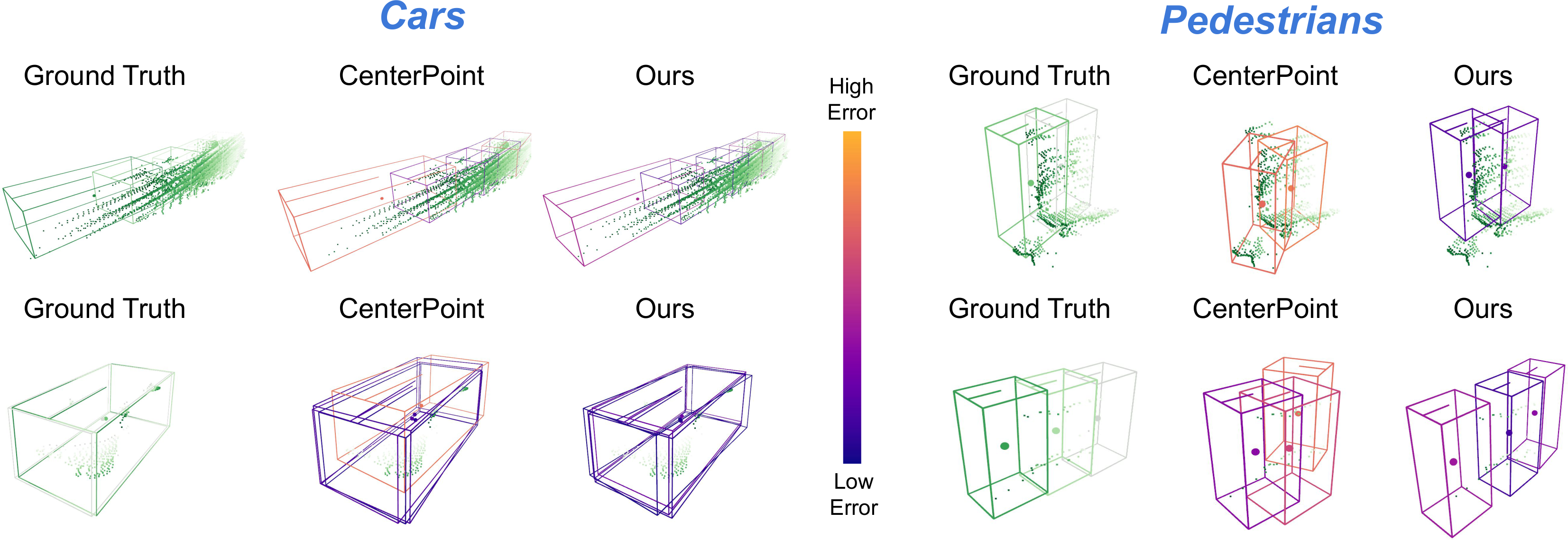}
    \caption{Qualitative results of our spatiotemporal sequence refinement. We visualize every 10th prediction for clarity. Predicted bounding boxes are colored according to their L2 center error. Refinement improves the temporal consistency of detections, especially for sparse sequences.}
    \label{fig:qual_refinement}
\end{figure}

In \cref{tab:waymo-tracking-results}, we compare SpOT to various tracking methods on the Waymo Open dataset. For pedestrians, SpOT significantly outperforms all previous methods. For vehicles, SpOT notably improves tracking over the CenterPoint baseline. Examining metric breakdowns (details in the supplementary), it becomes clear that SpOT is able to robustly track objects in more cluttered environments compared to previous methods. That is, we remove fewer detections in pre-processing to yield fewer \textit{Misses}, yet we still maintain a very low \textit{Mismatch} score. Additionally, we note that many contributions of the competitive method SimpleTrack \cite{simpletrack}, such as a 2-stage association strategy and a generalized-IoU similarity metric, could be seamlessly integrated into SpOT. 
The SSR module of SpOT exhibits greater improvements on the nuScenes dataset and on the pedestrian class in general. This is unsurprising because sparser LIDAR frames and smaller objects are expected to benefit disproportionately from increased temporal context. \cref{fig:qual_refinement} illustrates examples of our refined sequences compared to tracklets composed of off-the-shelf CenterPoint detections. We observe the greatest improvement in sequence quality when individual frames are sparse. Furthermore, we can qualitatively observe improved temporal consistency within sequences.

\begin{figure}[t]
\centering
    \includegraphics[width=0.68\textwidth]{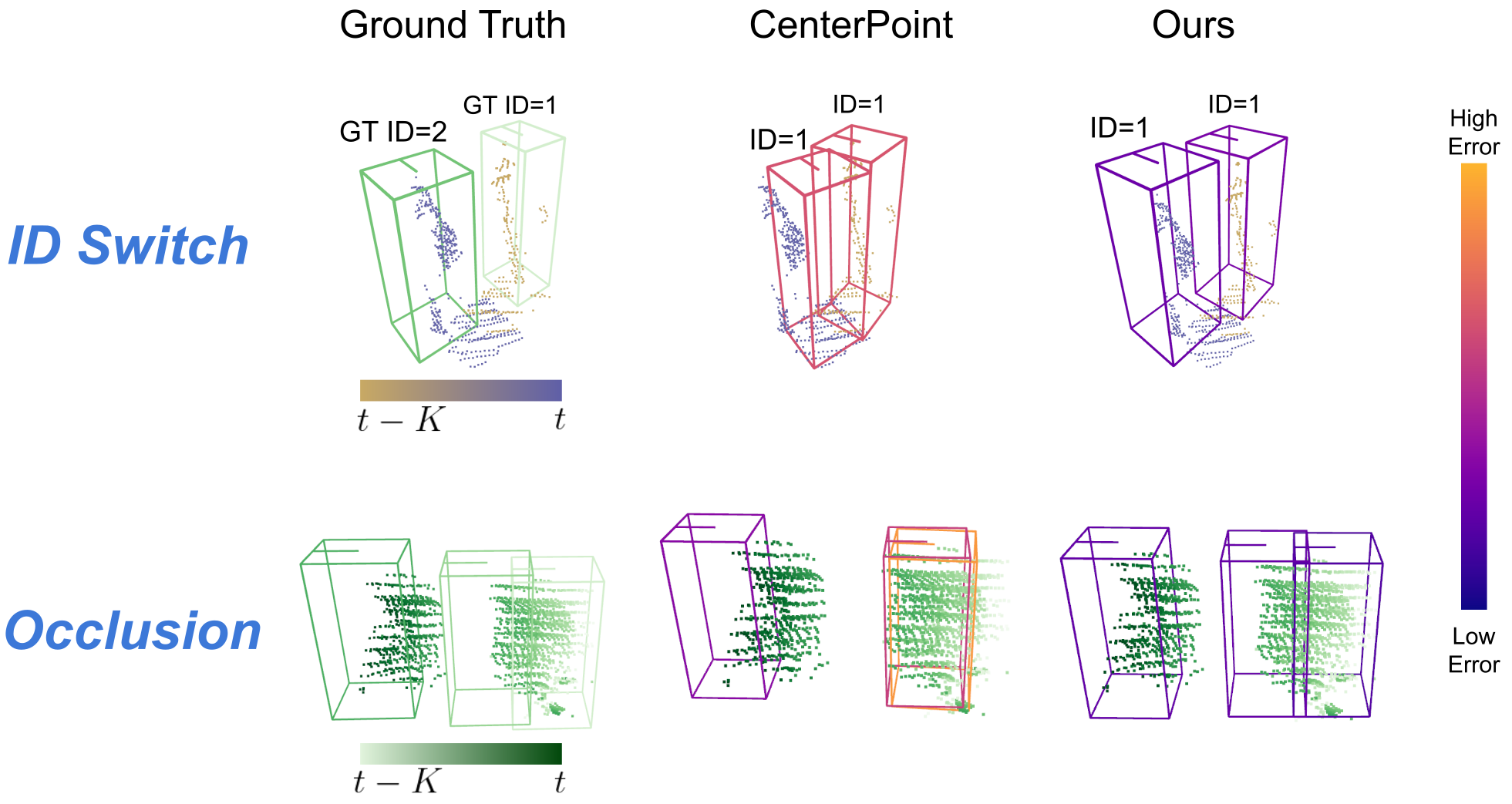}
   \caption{Examples of discontinuous object tracklets. We visualize every 10th prediction for clarity, and predicted boxes are colored according to L2 center error. Our refinement is robust to different types of input sequence discontinuities. In the first row, our refinement correctly updates bounding boxes to reflect the existence of two disjoint objects. In the second row, it correctly updates bounding boxes to reflect single-object continuity through occlusion. 
 }
 \vspace{-5mm}
\label{fig:incontiguous_seqs}
\end{figure}

Additionally, we observe that our SSR module can handle noisy input detections and/or associations by learning \textit{when} to make use of physical priors on object permanence and consistency. We provide some examples illustrating this property in \cref{fig:incontiguous_seqs}. The first row displays an example when both CenterPoint and SpOT encounter an ID-switch error in the tracklet. For CenterPoint, this error will be propagated to future prediction and association. For SpOT, even though we can not retroactively correct the misassociation, the SSR module still refines the sequence bounding boxes in a manner that accurately reflects two disjoint objects; this accurate update will help to avoid future tracking errors. The second row shows a discontinuous sequence due to \textit{occlusion} where different parts of an object is observed. Our SSR module refines the occluded region in a manner that reflects temporal continuity of a single object.

\subsection{Ablative Analysis}\label{sec:performance-analysis}

\paragraph{Length of Maximum History}
\cref{tab:sequence-length} reports how the length of maximum history affects tracking performance. The table emphasizes the significant advantage of using a large history. Because nuScenes is sampled at 20Hz and uses a sparse 32-beam LIDAR, we observe tracking performance monotonically improve up to a 40-frame history (2 seconds). In contrast, Waymo tracking performance peaks at a 10-frame history (1 second) and declines beyond 10 frames.

\paragraph{SSR Update Components}
Recall that for each object tracklet, our SSR module predicts per-timestep refinements consisting of a bounding box with velocity and a confidence score. \cref{tab:nuscenes-ablation} displays an ablation study on these two SSR refinements. All reported values are the AMOTA tracking metric on the nuScenes dataset. These results indicate that both bounding box and confidence refinements contribute to tracking, and we achieve the best performance when we refine both.

\paragraph{SSR Backbone Architecture}
\cref{tab:nuscenes-ablation}b displays an ablation analysis of our SSR backbone on the nuScenes dataset. All reported values are the AMOTA tracking metric. The first row shows tracking metrics with only a PointNet backbone. The second row corresponds to a two-branch backbone where the second branch consists of set abstraction and feature propagation. The third row corresponds to our full backbone. As observed, each part of our backbone improves sequence-to-sequence refinement.

\subsection{Runtime Analysis}\label{sec:runtime-analysis}
All components of SpOT except our SSR module are used and benchmarked in previous real-time tracking algorithms \cite{simpletrack,centerpoint,abmot3d}. We benchmark the real-time performance of our SSR module on an Nvidia RTX3090 GPU. On the nuScenes validation split, our SSR module averages at 51Hz per-frame for pedestrians and 28Hz per-frame for cars. On the Waymo validation split, it averages at 26Hz for pedestrian and 17Hz for vehicles.  

\begin{table}[t]
\centering
 \small
\begin{tabular}{ c c c @{\hskip 6pt} | @{\hskip 6pt} c c c}
\toprule
  \multicolumn{3}{c}{\textbf{nuScenes} (20Hz): AMOTA$\uparrow$} & \multicolumn{3}{c}{\textbf{Waymo} (10Hz): MOTA$\uparrow$} \\
  Num Sweeps \;\;& \textit{Car} \; & \textit{Pedestrian}  \; & Num Sweeps \;\; & \textit{Vehicle} \; & \textit{Pedestrian} \\
  \midrule
  10 & 84.7 & 81.1 & 1 & 53.4 & 58.4 \\
  20 & 85.0 & 81.4 & 5 & 54.6 & 59.8 \\
  30 & 85.0 & 81.7 & 10 & \textbf{55.7} & \textbf{60.4} \\
  40 & \textbf{85.1} & \textbf{82.5} & 15 & 54.5 & 60.2 \\
  \bottomrule
\end{tabular}
\vspace{3mm}
\caption{Ablation on how the length of maximum history (number of input LIDAR sweeps to SSR) affects the quality of tracking. Tracking performance peaks at a $2$ second history for nuScenes and a $1$ second history for Waymo.}
\vspace{-3mm}
\label{tab:sequence-length}
\end{table}
\begin{table}[t]
\centering
    \begin{subtable}[t]{0.5\textwidth}
        \centering
        \begin{tabular}{l @{\hskip 6pt} c @{\hskip 3pt} c}
            \multicolumn{3}{c}{\textbf{SSR Refinement Components}}\\
            \toprule
            Refinement Type & \textit{Car} & \textit{Pedestrian} \\
            \midrule
             No Refinement & 84.5 & 81.1 \\
             Boxes Only & 84.9 & 81.9 \\
             Confidences Only & 84.9 & 81.6\\
             Boxes and Confidences & \textbf{85.1} & \textbf{82.5} \\
            \bottomrule
        \end{tabular}
        \label{tab:autoregressive-ablate}
    \end{subtable}
    \hspace{4mm}
    \begin{subtable}[t]{0.45\textwidth}
        \centering
        \begin{tabular}{l @{\hskip 8pt} c @{\hskip 3pt} c}
            \multicolumn{3}{c}{\textbf{SSR Encoder}}\\ 
            \toprule
            Architecture & \textit{Car} & \textit{Pedestrian} \\
            \midrule
            PointNet only & 84.8 & 81.6 \\
            + SA and FP  & 84.8 & 81.7\\
            + Self-attention & \textbf{85.1} & \textbf{82.5}\\
            \bottomrule
        \end{tabular}
        \label{tab:backbone-ablate}
    \end{subtable}
    \vspace{3mm}
    \caption{Ablation experiments on the nuScenes dataset. All reported values are the AMOTA tracking metric. (a) Ablation holding out box and confidence-score refinements of our SSR module. (b) Ablation holding out parts of our refinement backbone. We denote set abstraction as SA and feature propagation as FP.}
    \label{tab:nuscenes-ablation}
\end{table}

\section{Discussion}

We have introduced SpOT, a method for 3D multi-object tracking in LIDAR data that leverages a spatiotemporal tracklet representation in the form of object bounding boxes and point cloud history. 
Furthermore, we have proposed a 4D refinement network to iteratively update stored object sequences after associating new detections at each frame.
Through evaluations on standard tracking benchmarks, SpOT compares favorably to prior works that use only single-frame tracks, thanks to the ability to leverage larger spatiotemporal context and use low-level geometry cues to improve bounding box and motion estimates.
Our method particularly excels given longer temporal history and when operating on pedestrians due to naturally increased occlusions and sparsity.

Though our results indicate a promising first step to improving 3D tracking with spatiotemporal representations, our approach does have limitations which hint at interesting future directions to explore, such as integrating the 2-stage association strategy and generalized-IoU similarity metric of~\cite{simpletrack}.

Furthermore, although we store object point cloud sequences and use this geometric data in refinement, we do not explicitly leverage the aggregated geometry in the track association step.
We believe further utilizing this shape context is an important future direction.

%
%
\clearpage
\bibliographystyle{splncs04}
\bibliography{egbib}
\end{document}


\pagestyle{headings}
\mainmatter
\def\ECCVSubNumber{2657}  

\title{Supplementary Material for ``Spatial-Temporal Refinement for 3D Object Tracking''} 

\titlerunning{Supplementary for ``Spatial-Temporal Refinement for 3D Object Tracking''}
\author{Colton Stearns\inst{1} \and Davis Rempe\inst{1} \and Jie Li\inst{2} \and Rares Ambrus \inst{2} \and Sergey Zakharov \inst{2} \and Vitor Guizilini \inst{2} \and Yanchao Yang \inst{1} \and Leonidas J. Guibas \inst{1}}
\authorrunning{C. Stearns et al.}
\institute{Stanford University \and Toyota Research Institute}
\maketitle

\section{Overview}
In this document, we provide additional implementation details, experimental analysis, qualitative results, and discussion. In \cref{sec:supp-architecture}, we provide further details of our encoding architecture. In \cref{sec:supp-implementation-details}, we discuss all implementation details of SpOT not covered in Sec. 4.3 of the main paper. In \cref{sec:supp-tracking-metrics}, we provide metrics definition for the two benchmarks. Finally, in \cref{sec:supp-results}, we provide more fine-grain discussion of our method with qualitative results as well as a supplemental ablation study.

\section{Additional Architecture Details}\label{sec:supp-architecture}
\subsubsection{Split Self-Attention Positional Encoding}
We refer the reader to Sec. 3.2 of the main paper for an overview of the split self-attention encoder used by our SSR module. We highlight that our positional encoding differs from previous works that utilize self attention \cite{3detr}. First, our positional encoding does not utilize a fourier coordinate transformation, \textit{i.e.} there is no $[\sin(x), \cos(x)]$ transformation. Second, we \textit{concatenate}, instead of add, the positional encoding to the anchor features. Experimentally, we find these modifications improve training in our novel 4D setting. 

\paragraph{Network Loss Hyperparameters}
Sec. 3.2 of the main paper provides an overview of our network's training losses. We set our network loss weights as follows: $w_{\text{c}}=3.0$, $w_{\theta}=3.0$, $w_{\text{vel}} = 1.5$, $w_{\text{wlh-cls}} = 1.0$, $w_{\text{wlh-res}}=1.5$, and $w_{\text{conf}}=1.0$.  We set the confidence-loss temperature to $\alpha=0.75$ for nuScenes cars, $1.0$ for nuScenes pedestrians, $1.2$ for Waymo vehicles, and $2.4$ for Waymo pedestrians.

\section{Additional Implementation Details}\label{sec:supp-implementation-details}

\subsection{Training-Time Augmentations}
\paragraph{Iterative Sequence Refinement}
During training, we stochastically update each batch of training sequences multiple times, \textit{i.e.} the network sees its own output as input. Concretely, for input training sequence $\mathbf{\Bar{T}}_t$, we apply our SSR module to generate a refined training tracklet: $\text{\textbf{SSR}}(\mathbf{\Bar{T}}_t) = \mathbf{\Bar{T}'}_t$. With probability $p_{\text{end}}$, we end the refinement and assign our output $\mathbf{T}_t = \mathbf{\Bar{T}'}_t$. Otherwise, we set $\mathbf{\Bar{T}}_t \leftarrow \mathbf{\Bar{T}'}_t$ and repeat. We limit the maximum number of iterative refinements to 4, and we set $p_{\text{end}} = \min(1 - \frac{\text{EPOCH}}{8}, 0.4)$. We find this iterative strategy noticeably improves training on the Waymo Open dataset. On the nuScenes dataset, we observe little improvement and ultimately leave it out to improve training efficiency.

\paragraph{Training Augmentation}
We apply four training sequence augmentations. First, we uniform-randomly drop the leading [1,K] frames of the sequence. Second, we apply a uniform-random rotation, scaling, and reflection to all tracklet bounding boxes and points; we sample rotation between $[-1.57, 1.57]$ radians, sample scaling between $[-5, 5]$ percent, and reflect about the x-axis with probability $0.5$. Third, we apply a \textit{single} uniform-random rotation, scaling, and translation to \textit{all} tracklet bounding boxes; we sample translation between $[-0.2, 0.2]$ meters, rotation between $[-0.25, 0.25]$ radians, and scaling between $[-10, 10]$ percent. Finally, we  apply \textit{per-frame} uniform-random translations and rotations to each tracklet bounding box; we sample translations between $[-0.1, 0.1]$ meters and rotations between $[-0.1, 0.1]$ radians. We use the same augmentations for all object classes.

\subsection{Training Schedule}
During training, we use the Adam optimizer \cite{adam-optimizer} with an exponentially decaying learning rate. We set our initial learning rate to $0.0025$ and our decay rate to $0.95$ per epoch. We train in parallel across 4 Nvidia A100 GPUs and use a global batch size of 300 sequences. We finish training after 10 epochs for pedestrians and 20 epochs for cars/vehicles.

\section{Tracking Metrics}\label{sec:supp-tracking-metrics}
\subsection{MOTA and MOTP}
The Waymo Open Dataset \cite{waymo-open} evaluates tracking using the MOTA and MOTP metrics \cite{clear-mot}. Multiple Object Tracking Accuracy (MOTA) is defined as:
\begin{equation}
\text{MOTA}=1-\frac{\sum_{t}\left(\text{MISS}_{t}+ \text{FP}_{t}+ \text{MISMATCH}_{t}\right)}{\text{GT}}
\end{equation}
where $\text{MISS}_t$, $\text{FP}_t$, and $\text{MISMATCH}_t$ respectively denote the number of \textit{missed} tracklets, \textit{false positive} tracklets, and \textit{mismatches} at time $t$. $\text{GT}$ denotes the number of all ground-truth tracklets. A mismatch (also denoted identity-switch) occurs when a current tracklet is assigned to a ground-truth object that differs from its previous ground-truth assignment. Thus, MOTA can be decomposed into three equivalent parts: (1) identifying all objects in the frame, (2) not identifying false-positives, and (3) consistently re-identifying objects between frames.  

Multiple Object Tracking Precision (MOTP) is defined as:
\begin{equation}
    \text{MOTP} =\frac{\sum_{i, t} d_{i, t}}{\sum_{t} \text{TP}_{t}}
\end{equation}
where $d_{i,t}$ denotes the L2 center error of the $i$'th true-positive tracklet at time $t$, and $\sum_{t} \text{TP}_{t}$ denotes the total number of true-positive tracklets. Thus, MOTP conveys the quality of center estimates for all correctly predicted object tracklets.

\subsection{AMOTA and AMOTP}
The nuScenes dataset \cite{nuscenes} evaluates tracking using the AMOTA and AMOTP metrics \cite{abmot3d}. AMOTA and AMOTP address the issue that the highest achievable MOTA often occurs at a low recall; that is, maximizing MOTA often causes tracking methods to \textit{remove} low confidence detections due to their causing an abundance of false-positives and mismatches. Concretely, Average Multiple Object Tracking Accuracy (AMOTA) averages a recall-weighted MOTA over $n$ evenly-spaced recall thresholds:
\begin{equation}
    \text {AMOTA} = \frac{1}{n-1} \sum_{r \in\left\{\frac{1}{n-1}, \frac{2}{n-1} \ldots 1\right\}} \text {MOTAR}
\end{equation}
For a given recall threshold, $r$, the recall-weighted MOTA metric, MOTAR, is defined as:
\begin{equation}
    \operatorname{MOTAR}=\max \left(0,1-\frac{\text{MISS}_{r} + \text{FP}_{r} + \text{MISMATCH}_{r} -(1-r) * \text{GT}}{r * \text{GT}}\right)
\end{equation}
where $\text{MISS}_r$, $\text{FP}_r$, and $\text{MISMATCH}_r$ respectively denote the number of missed tracklets, false positive tracklets, and mismatches over \textit{all} times for a recall threshold $r$. GT denotes the total number of ground-truth tracklets.

Average Multiple Object Tracking Precision (AMOTP) averages MOTP over all recall thresholds, \textit{i.e.}:
\begin{equation}
    \text {AMOTP} =\frac{1}{n-1} \sum_{r \in\left\{\frac{1}{n-1}, \frac{2}{n-1}, . ., 1\right\}} \text{MOTP}_r
\end{equation}

\subsection{Discussion}
While evaluating on AMOTA reflects the \textit{average} MOTA over all recall thresholds, evaluating on MOTA incites selection of the \textit{maximum} MOTA over all recall thresholds. Although correlated, modern tracking algorithms often encounter a substantial tradeoff between the two metrics. For instance, greedy and center-distance association strategies have been shown to be more effective for AMOTA \cite{centerpoint,simpletrack}. Hungarian-matching and Intersection-Over-Union are more effective for MOTA \cite{abmot3d,simpletrack}. We highlight this as a concern in Lidar 3D multi-object tracking: methods often only evaluate one of these metrics and offer no analysis of the other. Contrary to this trend, we showcase SpOT's robustness across both metrics via our evaluation on the nuScenes (AMOTA) and Waymo Open (MOTA) datasets.

\section{Additional Analysis}\label{sec:supp-results}
Please refer to Sec. 4 of the main paper for a comprehensive reporting of SpOT's tracking performance and an extensive ablation study of SpOT's design choices.

\begin{table}[t]
    \centering
    \begin{tabular}{l @{\hskip 12pt} c @{\hskip 9pt} c @{\hskip 9pt} c @{\hskip 9pt} c}
        \toprule
         Method / Pedestrian & MOTA$\uparrow$ & FP\%$\downarrow$ & Miss\%$\downarrow$ & Mismatch\%$\downarrow$ \\
         \midrule
         &  \multicolumn{4}{c}{\textit{Tracklet Birth Threshold 0.75}}\\
         CenterPoint \cite{centerpoint} & 54.9 & 10.0 & 34.0 & 1.13 \\
         SpOT-No-SSR (Ours) & 55.8 & 10.5 & 33.3 & 0.38 \\
         SpOT (Ours) & \textbf{60.4} & \textbf{9.5} & \textbf{29.8} & \textbf{0.34} \\
         \midrule
          & \multicolumn{4}{c}{\textit{Tracklet Birth Threshold 0.60}} \\
          CenterPoint \cite{centerpoint} & 51.1 & 9.8 & 35.2 & 3.80 \\
          SpOT-No-SSR (Ours) & 56.5 & 11.4 & 31.5 & 0.61 \\
          SpOT (Ours) & \textbf{60.5} & \textbf{11.3} & \textbf{27.6} & \textbf{0.56} \\
         \bottomrule
    \end{tabular}
    \vspace{3mm}
    \caption{Tracking performance on the pedestrian class of the Waymo Open dataset validation split. We compare SpOT and CenterPoint with controlled tracklet birth thresholds. Best result in each threshold is bolded.}
    \vspace{-5mm}
    \label{tab:supp-waymo-tracking-details}
\end{table}

\subsection{Waymo Tracking Analysis}
Tab. 2 of the main paper reports the tracking results of SpOT on the Waymo Open dataset comparing the state-of-the-arts.
In our experiments on the Waymo dataset, we found that tracking performance of some state-of-the-art algorithms tends to be sensitive to the tracklet birth confidence threshold, $c_{thresh}$, which determines when an unmatched detection will become a tracklet (e.g. a lower $c_{thresh}$ allows more unmatched detections to become tracklets). 
This is not suprising as MOTA focuses more on the high-confidence region of tracking results. For a fair comparison, in Tab. 2 we report results with the optimized $c_{thresh}$ value for each tracking algorithm. 

In \cref{tab:supp-waymo-tracking-details}, we provide ablation analysis on $c_{thresh}$ and show our robustness towards this parameter. We evaluate performance with the original threshold used in CenterPoint, $c_{thresh} = 0.75$, as well as a lower threshold, $c_{thresh} = 0.60$. The lower threshold creates a more challenging setting as more unmatched detections will be treated as tracklets; this creates a more cluttered environment during association. 
As shown in \cref{tab:supp-waymo-tracking-details}, our method is able to provide comparable results across both thresholds while CenterPoint's performance is negatively affected by the lower threshold.
This example showcases SpOT's robustness against cluttered scenes thanks to the use of dense spatiotemporal information.

\subsection{nuScenes Tracking Analysis}
\begin{figure}[t]
\centering
    \vspace{-5mm}
    \includegraphics[width=0.5\textwidth]{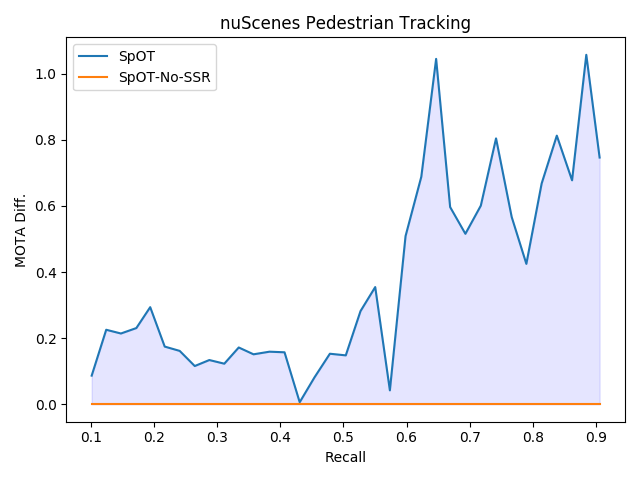}
    \caption{A detailed analysis of pedestrian tracking on the nuScenes dataset. Shown is the difference in MOTA between SpOT and the SpOT-No-SSR baseline for tracklet recalls thresholds in [0.10, 0.91].}
    \label{fig:nuscenes-ped-amota}
\end{figure}
\begin{figure}[b!]
    \centering
    \includegraphics[width=0.65\textwidth]{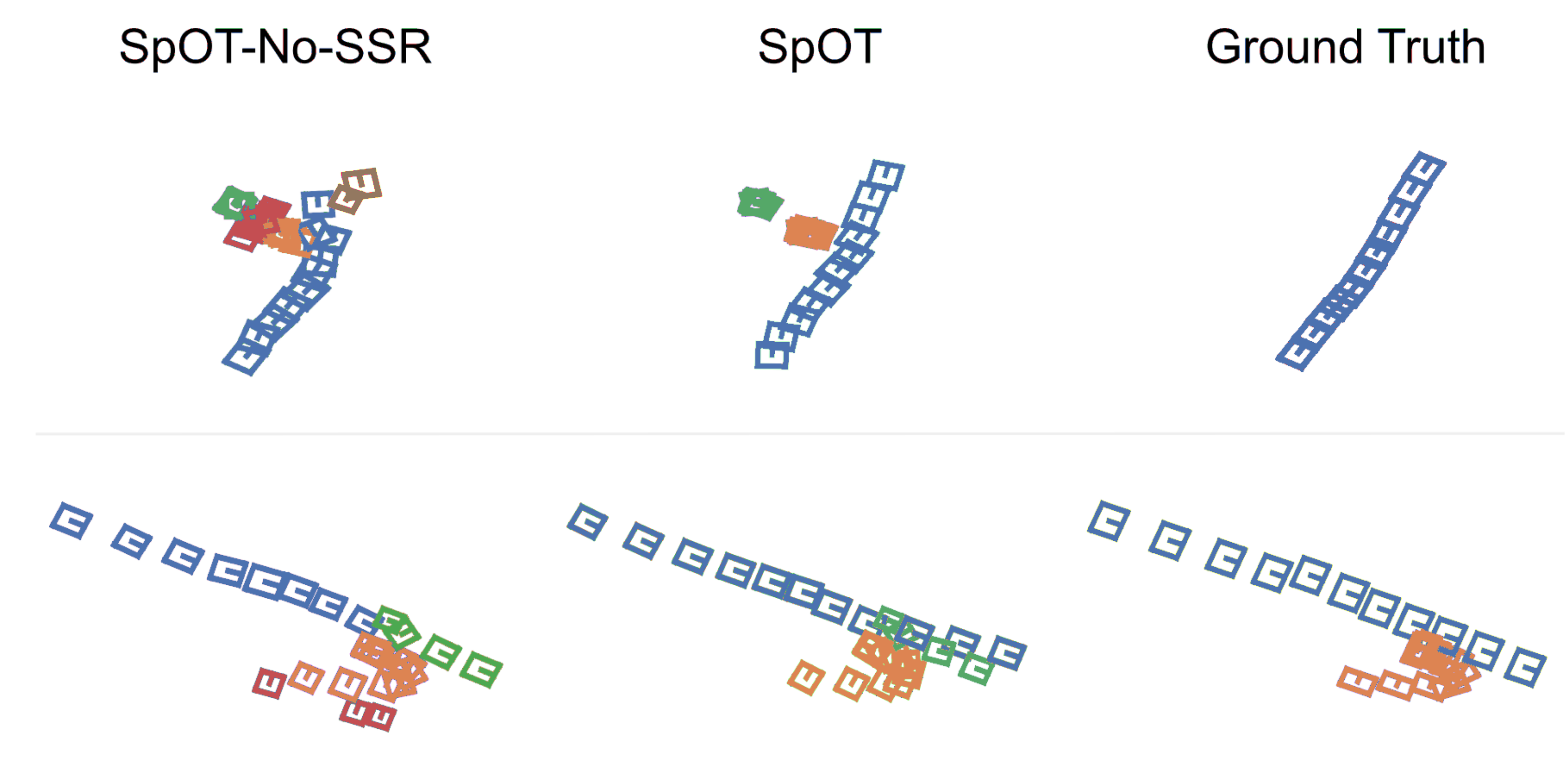}
    \caption{Two example birds-eye-view visualizations of pedestrians tracking over many frames on the nuScenes dataset. Tracking predictions are colored consistently. SpOT-No-SSR shuffles tracklets, resulting in mismatches and additional false-positives. In contrast, SpOT establishes cleaner sequences via its bounding-box refinement.}
    \vspace{-3mm}
    \label{fig:nusc-bev-box-refine}
    
\end{figure}

In Tab. 1 of the main paper, we report the final AMOTA tracking metric of SpOT on the nuScenes dataset.
In this section, we offer more fine-grain analysis on nuScenes to better analyze the behavior of our proposed algorithm.

In \cref{fig:nuscenes-ped-amota}, we visualize the \textit{difference} in the MOTA metric with respect to the SpOT-No-SSR baseline at different recall thresholds.
As depicted in \cref{fig:nuscenes-ped-amota}, SpOT consistently improves the tracking results at different recall levels.
It's also worth noticing that SpOT improves MOTA disproportionately at higher recall thresholds.
This observation furthers the claim that SpOT is robust in cluttered scenes due to the use of dense spatiotemporal information, which is consistent with what we observe in \cref{tab:supp-waymo-tracking-details}.

In addition, we also provide some qualitative examples showcasing SpOT's improvements in individual tracking scenarios. 

In \cref{fig:nusc-bev-box-refine}, we provide two illustrative examples of how SpOT's bounding-box refinement improves tracking compared to the SpOT-No-SSR baseline. In the SpOT-No-SSR column of both examples, we observe that poor motion estimates and poor sequence continuity cause tracklet fragmentation and mis-association. In contrast, due to the sequence-to-sequnce refinement, SpOT avoids fragmentation and establishes more accurate tracklets. 
\begin{figure}[t]
    \centering
    \includegraphics[width=1.0\textwidth]{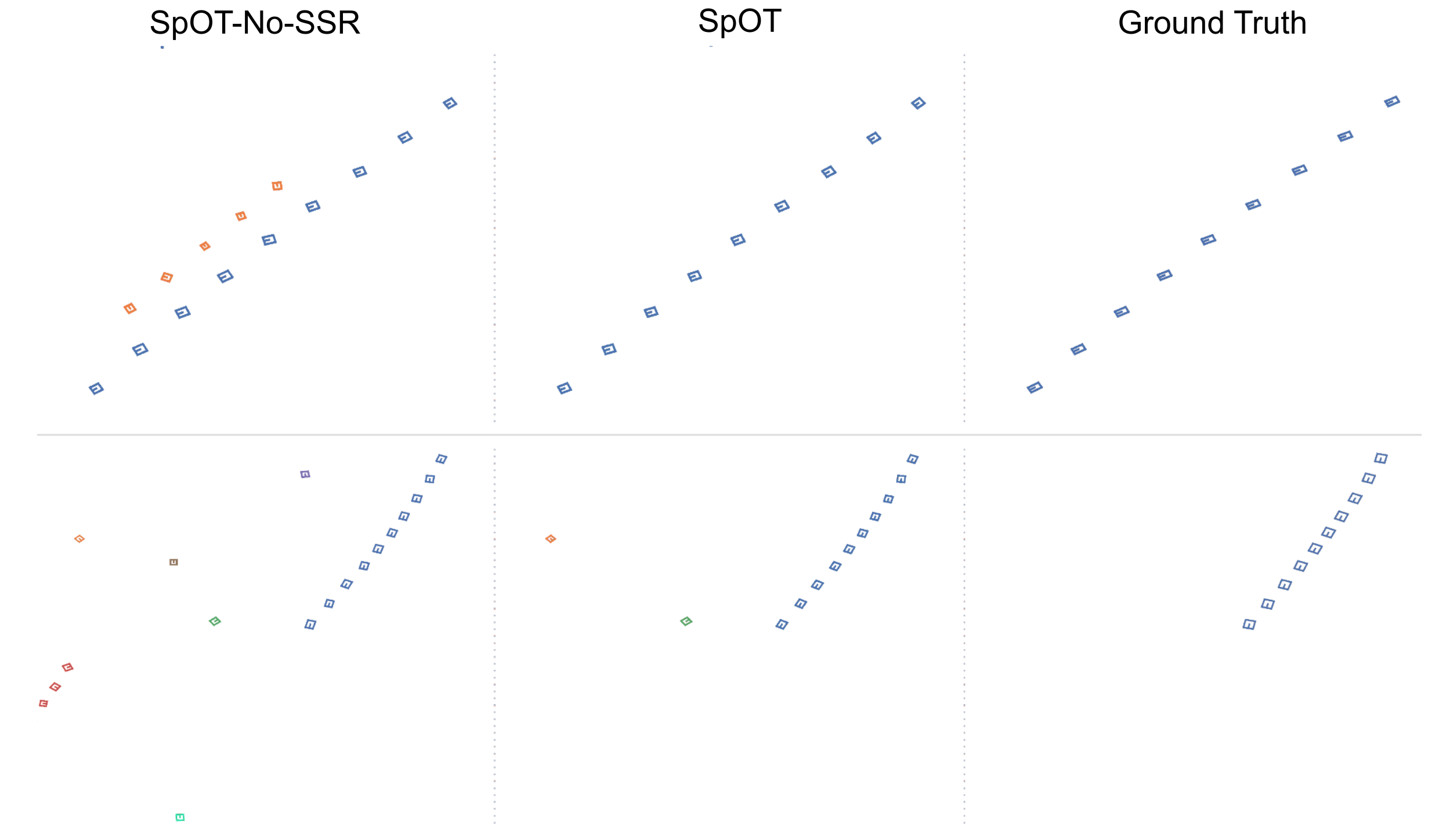}
    \caption{Two example birds-eye-view visualizations of pedestrians tracking over many frames on the nuScenes dataset. Tracking predictions are colored consistently. For the visualized recall threshold of 91.1\%, SpOT's confidence refinement successfully identifies and removes false-positive tracklets.}
    \vspace{-3mm}
    \label{fig:nusc-bev-conf-refine}
\end{figure}

In \cref{fig:nusc-bev-conf-refine}, we provide two illustrative examples of how SpOT's confidence refinement reduces the number of false-positive tracklets. In the SpOT-No-SSR column of both examples, we observe many false-positive tracklets, \textit{i.e.} tracklets with confidence-scores that lie within the visualized recall threshold of 91.1\%. After updating tracklet confidence-scores with its sequence-to-sequence refinement, SpOT is able to remove many false-positive tracklets.

%
%
\clearpage
\bibliographystyle{splncs04}
\bibliography{egbib}